\documentclass[11pt,a4paper]{article}
\usepackage[T1]{fontenc}
\usepackage[utf8]{inputenc}

\usepackage{url}
\usepackage{graphicx}
\usepackage{multirow}
\usepackage{amssymb}
\usepackage{scrextend}
\usepackage{flushend}
\let\mfs\multiplefootnoteseparator
\author{
  Michael Fire\\
  Sizmek Technologies LTD \\
  \texttt{michael.fire@sizmek.com}
  \and
  Jonathan Schler\\
  Sizmek Technologies LTD \\
  \texttt{jonathan.schler@sizmek.com}
}
\title{Exploring Online Ad Images Using a Deep Convolutional Neural Network Approach}

\date{}

\begin{document}

\maketitle
\begin{abstract}
Online advertising is a huge, rapidly growing advertising market in today's world. One common form of online advertising is using image ads. A decision is made (often in real time) every time a user sees an ad, and the advertiser is eager to determine the best ad to display. Consequently, many algorithms have been developed that calculate the optimal ad to show to the current user at the present time. Typically, these algorithms focus on variations of the ad, optimizing among different properties such as background color, image size, or set of images.
However, there is a more fundamental layer. Our study looks at new qualities of ads that can be determined before an ad is shown (rather than online optimization) and defines which ads are most likely to be successful. 

We present a set of novel algorithms that utilize deep-learning image processing, machine learning, and graph theory to investigate online advertising and to construct prediction models which can foresee an image ad's success. 
We evaluated our algorithms on a dataset with over 260,000 ad images, as well as a smaller dataset specifically related to the automotive industry, and we succeeded in constructing regression models for ad image click rate prediction. 
The obtained results emphasize the great potential of using deep-learning algorithms to effectively and efficiently analyze image ads and to create better and more innovative online ads. Moreover, the algorithms presented in this paper can help predict ad success and can be applied to analyze other large-scale image corpora.
\end{abstract}

\section{Introduction}
Online advertising is one of the largest advertising markets in the world, and it has grown rapidly in recent years~\cite{iab2014}. According to ComScore, about 5.3 trillion display ads were delivered in the U.S. throughout 2012~\cite{radwanick2012us}. 
Furthermore, Magna Global~\cite{magna2014} predicts that online advertising will outgrow TV advertising, which is currently the leading advertising medium in the U.S., and by 2017 online ad revenues will reach 72 billion dollars. Additionally, Forrester forecasts that by 2019 U.S. advertisers will spend over 100 billion dollars on digital advertising, while estimating TV advertising to be only 90 billion dollars~\cite{theeara2014}. 

One form of online advertising is web banners (also referred to as banner ads) in which ads are embedded into web pages  as static images. The web banners seek to attract traffic to  advertisers' websites by prompting website visitors to engage with the ads, mostly by clicking the ads and then being directed to the advertisers' websites. A prevalent method to measure the success of a web banner is measuring the click-through rate (CTR) of an ad by calculating the ratio of the number of times an ad was clicked to the number of times an ad was presented. Moreover, one of the most common advertising revenue models is paying according to ad performance with cost-per-click (CPC) billing~\cite{richardson2007predicting}. Therefore, predicting an online ad's success becomes 
fertile ground for research~\cite{richardson2007predicting,graepel2010web,mcmahan2013ad}. 

In this study, we used a large-scale, unique-images dataset, which consisted of 261,752 banner ads from 23 categories, to understand better the world of online advertising. To explore this dataset, we utilized deep-learning algorithms to explore and analyze this dataset in the following manner:

First, we used a trained deep convolutional neural network~\cite{fukushima1980neocognitron} to identify objects that appeared in each ad.  Afterwards, we used the identified objects in each image along with graph theory algorithms to understand the connections among the different ad categories.

By identifying objects that appeared in each ad, we can gain some interesting insights regarding ad categories. For example, we can notice that many ads under the Telecom category contain traffic lights, while many ads under the Gaming category contain space shuttles and pay phones. 
By recognizing which objects appear under each category, we can better understand the visualization characteristics of image ads in general, and characteristics of image ads in specific categories in particular. Greater understanding promotes improvement and innovation. This type of information can give advertisers recommendations on which objects to embed in their ads in order to make them more appealing, effective, and lucrative.

Second, we used the pretrained deep convolutional neural network to transfer each ad image to its representative  vector. Then, we used unsupervised clustering algorithms to divide the ad images into disjointed clusters, which we explored to gain further insights. Using this method, we could determine the main ad banner types that existed in the image corpus.

Lastly, we drilled down to explore web banners that are related to the automotive industry by analyzing an additional image dataset with 34,451 image ads connected with the automotive industry. To inspect this dataset, we transferred each ad into its corresponding vector and divided them into disjoint clusters for exploration. We then utilized the calculated ad vector representations to create regression models which were able to predict each ad's CTR. 

Throughout this study, we demonstrate the value of deep-learning image processing algorithms in better understanding the domain of image ads. Our methodology provides new insights into this burgeoning field, as well as offering analysis techniques that can be used to reveal significant patterns in other large-scale image corpora. Moreover, these methods can lead to important resources for advertisers wanting to present their products in the most innovative, effective manner.

\subsection{Contributions}
To our knowledge, this study is the first to offer algorithms to analyze a large-scale corpus of image ads by utilizing 
deep-learning image processing algorithms.
Our key contributions presented in this paper are as follows: 
\begin{itemize}
	\item novel techniques to analyze a large-scale categorized image corpus.
	\item algorithms to infer connections among image categories.
	\item algorithms for constructing prediction models for ad image CTRs.
\end{itemize}

\subsection{Organization}
The remainder of the paper is organized as follows: In Section~\ref{sec:related}, we provide an overview of various related studies. In Section~\ref{sec:methods}, we describe the methods, algorithms, and experiments used throughout this study. 
Next, in Section~\ref{sec:results}, we present the results of our study. 
Then, in Section~\ref{sec:diss}, we discuss the obtained results. Lastly, in Section~\ref{sec:conclusions}, we present
our conclusions from this study and also offer future research directions.

\section{Background and Related Work}
\label{sec:related}
In this study, we primarily utilize three types of algorithms: (a) \textit{deep convolutional neural networks} for processing ad images; (b) \textit{clustering algorithms}  for separating the ad images into clusters and  understanding the connections between the various ad categories; and (c) \textit{supervised machine-learning algorithms} for predicting an image ad's CTR. 
In the rest of this section, we present a brief overview on each one of these types of algorithms.

\subsection{Deep-Learning  Algorithms for Image Recognition }
Deep learning is a new area of machine learning in which a set of algorithms attempts to model high-level abstraction in data. 
One of the common applications of deep-learning algorithms is image processing. 
By utilizing deep-learning algorithms to process images, researcher have recognized  hand-written digits~\cite{ciresan2012multi}, identified traffic signs~\cite{ciresan2012multi}, detected facial keypoints~\cite{sun2013deep}, classified objects in images~\cite{krizhevsky2012imagenet}, and more.

For object classification and image categorization there are well-known public datasets, such Caltech-101~\cite{fei2007learning}, CIFAR-10~\cite{krizhevsky2010convolutional}, and MNIST~\cite{lecun1998mnist}, which can be used as benchmarks to evaluate new and existing image processing algorithms.  In recent years, deep-learning algorithms have achieved state-of-the-art results on many of these datasets.\footnote{\url{http://rodrigob.github.io/are_we_there_yet/build/classification_datasets_results.html}}
One of the most popular image processing challenges is the ImageNet Large Scale Visual Recognition Challenge (ILSVRC). The ILSVRC Challenge has run annually since 2010 and has attracted  participation from more than fifty institutions. One of the two categories of the ILSVRC Challenge is predicting if  an object, out of 1,000 predefined object classes, exists or does not exist in an image~\cite{russakovsky2014imagenet}. In 2012, Krizhevsky et al.~\cite{krizhevsky2012imagenet} used a deep convolutional neural network to classify the 1.3 million high-resolution images in the ILSVRC-2010 training set into the 1,000 different classes. Their trained classifier achieved a top-1 error rate of 39.7\% and top-5 of  18.9\%.
Additionally, Krizhevsky et al. achieved an outstanding top-5 test error rate of 15.3\% on the ILSVRC-2012 dataset.
In 2014, Szegedy et al., also known as the GoogLeNet team, utilized a convolutional neural network to win first place at 
the ILSVRC-2014 classification challenge, with a top-5 test error of 6.67\%~\cite{szegedy2014going}. 
Recently, He et al.~\cite{he2015delving} from Microsoft Research achieved a 4.94\% top-5 test error rate on the  ImageNet  2012  classification dataset. According to He et al.,
their classifier demonstrated unprecedented results that were ``surpassing human-level performance on ImageNet classification.''  

There are many deep-learning software frameworks, such  Theano~\cite{bergstra2011theano}, Caffe~\cite{jia2014caffe}, Deeplearning4j,\footnote{\url{http://deeplearning4j.org/}} and GraphLab~\cite{low2014graphlab},  that enable researchers to easily run and evaluate deep-learning algorithms. In this study, we chose to utilize 
Graph-Lab's implementation of an image category classifier~\cite{gu2014}  that was derived from the study of Krizhevsky et al.~\cite{krizhevsky2012imagenet}. Throughout this study, we use the pretrained deep-learning classifier~\cite{gu2014} to predict objects that appear in ad images. Moreover, we used the classifier to transfer images to their vector representations.

It is worth mentioning that although deep-learning algorithms, such as deep convolutional neural networks, are very useful and present state-of-the-art results in many image categorization and object classification challenges, these types of algorithms have flaws that need to be kept in mind~\cite{szegedy2013intriguing}.  For example, Artem Khurshudov recently demonstrated that many implementations of deep convolutional neural network classifiers classified  a leopard print sofa image as Felinae images~\cite{khurshudov2015}.

\subsection{Clustering Algorithms}
During this study, we used two types of clustering algorithms in order achieve two separate goals. 
To understand the connections among different ad categories and objects (see Section~\ref{method:all_ads}) we utilized a community detection algorithm. This type of algorithm organizes graph vertices into communities. Usually, many links exist among vertices in the same community, while comparatively fewer links exist among vertices in different communities. 
There are many various community detection algorithms~\cite{fortunato2010community}. In this study we chose to use
the GLay~\cite{su2010glay} clustering algorithm that is implemented as part of Cytoscape's clusterMaker2 application.\footnote{\url{http://apps.cytoscape.org/apps/clusterMaker2}}

Our second type of clustering algorithm was used to identify clusters of images according to image properties (see Section~\ref{method:all_ads}). For this. we used the k-means clustering algorithm~\cite{arthur2007k}.  K-means is a widely used  clustering algorithm which separates $n$ observations into $k$ clusters by seeking to minimize the average squared distance between points in the same cluster. To use the k-means on a dataset one needs to preselect the number of clusters $k$. There are various algorithms to identify the recommended number of clusters, such as the gap statistic method that was presented by Tibshirani et al.~\cite{tibshirani2001estimating}.

In this study, we used the k-means++ algorithm, which augments k-means with a simple, randomized seeding technique that can quite dramatically improve both the speed and the accuracy of k-means clustering~\cite{arthur2007k}. Additionally, we chose the number of clusters $k$ using a simple heuristic that is described in detail in Section~\ref{method:all_ads}.

\subsection{Ad Success Prediction}
Displaying the right online ads that will be clicked by a user can greatly influence both the user's experience and the revenue from ads of advertisers that a use cost-per-click billing model. 
Therefore, in the last decade, predicting if a user will click an  online ad has become fertile ground for researchers.

In 2006, Regelson and Fain~\cite{regelson2006predicting} introduced a method for improving CTR prediction accuracy using keyword clusters.
In 2007, Richardson et al.~\cite{richardson2007predicting} utilized various features of search ads (i.e., ads that appear  mainly in a search engine's results) to construct a  logistic regression model that can predict the click-through rate of new ads. In 2008, Dembczy\'{n}ski et al.~\cite{dembczynski2008predicting} utilized the Beyond Search dataset to construct their model for CTR prediction using an ensemble of decision rules. Moreover, Dembczy\'{n}ski et al. demonstrated how their suggested algorithms can be used to provide recommendations in order to improve the ads' quality.
Later in 2011, Wang et al.~\cite{wang2011} attempted to predict the ideal number of ads that should be displayed for a given search-engine query.

Recently, the following competitions made public large-scale, CTR-related datasets:
\begin{itemize}
\item \textit{2012 KDD Cup},\footnote{\url{http://www.kddcup2012.org/}}  
 in which the challenge was to predict the CTR of ads in online social networks given the query and user information.
 \item \textit{CriteoLabs Display Advertising Challenge},\footnote{\url{https://www.kaggle.com/c/criteo-display-ad-challenge}  \label{kaggle1}} in which the challenge was to create accurate algorithms for CTR estimation on Criteo's dataset that contains a portion of Criteo's traffic over a period of 7 days.
\item \textit{Avazu's Click-Through Rate Prediction Challenge},\footnote{\url{https://www.kaggle.com/c/avazu-ctr-prediction} \label{kaggle2}} in which the challenge was to predict, using Avazu's released dataset that contains 10 days of click-through data,   whether a mobile ad will be clicked or not.

\end{itemize}
With the release of these datasets, there was an immense amount of interest, over a thousand researchers according to the competition pages on Kaggle,\footref{kaggle1}\mfs\footref{kaggle2} in developing CTR prediction models. In both the CriteoLabs Display Advertising Challenge and Avazu's Click-Through Rate Prediction Challenge, the winners used field-aware factorization machines (FFM) to achieve the best overall results~\cite{libffm2015}.

\section{Methods and Experiments}
\label{sec:methods}

\begin{figure}
\centering
\includegraphics[scale=0.55]{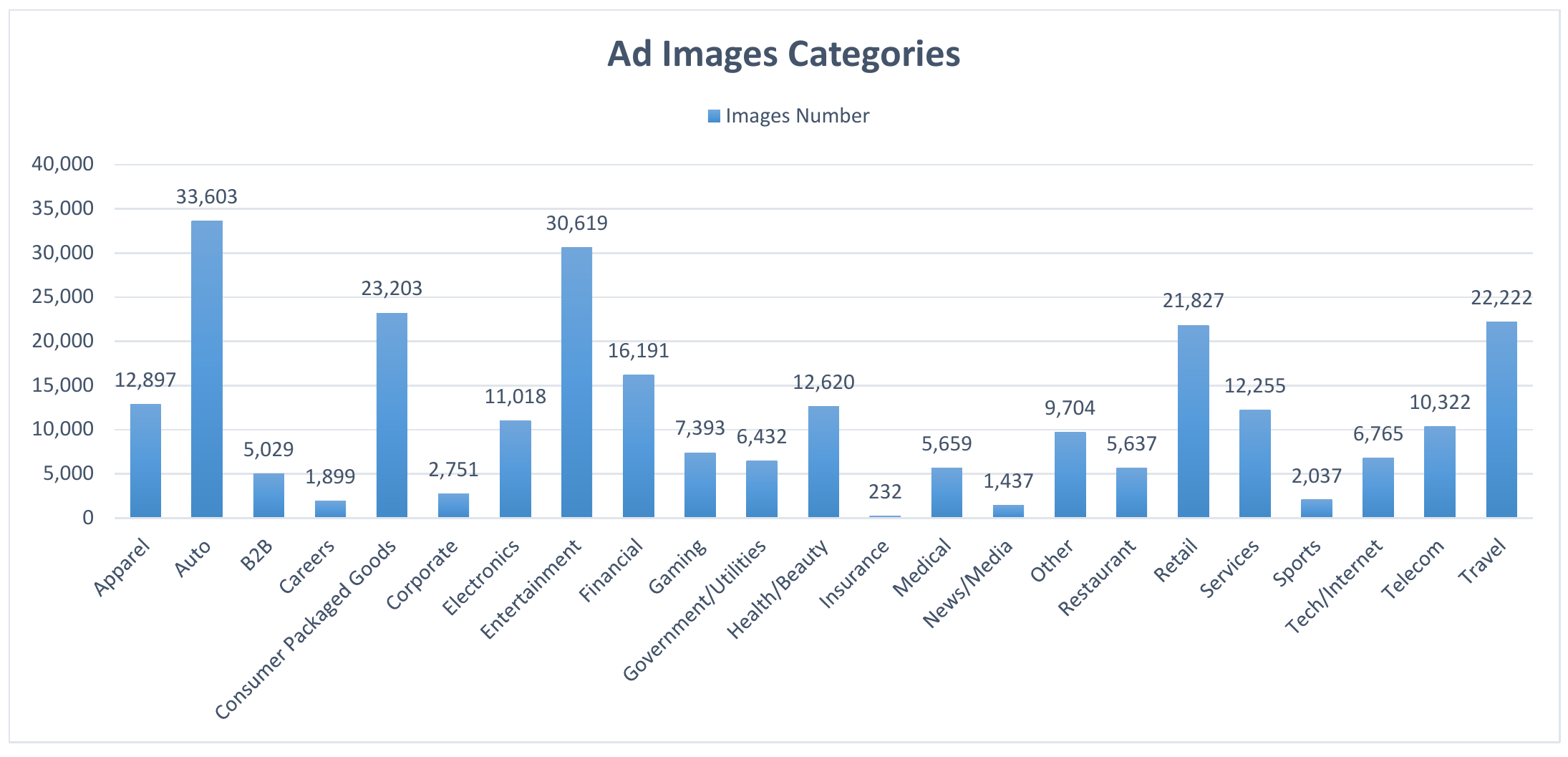}
\caption{Ad images distributed into 23 categories.}
\label{fig:categories_stat}
\end{figure}

In this study, we utilized two ad image datasets. The first dataset, referred to as  \textit{all-ads dataset}, contains 261,752  images\footnote{Each image in our dataset has its own unique MD5 hash. However, it is worth noticing that the same ad can still appear multiple times with minor changes, such as same images with different resolutions, different file formats,  different colors, or even very minor changes in the ads' texts.} collected from advertising campaigns of over 6,500 brands.\footnote{Brand can be a company, or a specific product of a company.} These images can be divided into  23 unique categories (see Figure~\ref{fig:categories_stat}).
The second dataset, referred to as the \textit{auto-ads dataset}, contains 34,451 ad images\footnote{The image ads in the auto-ads dataset were created in a separate process than the ad images in the all-ads dataset. Nevertheless, both datasets share 25,611 ad images with unique MD5 hash.} from over 800 brands that were labeled as ads which are related only to the automotive industry.

\subsection{All-Ads Dataset}
\label{method:all_ads} 
To better understand the all-ads dataset, we chose to utilize the ImageNet based deep-learning classifier (referred to as the ImageNet classifier) that can classify each image into 1,000 different classes according to the objects that appear in each image~\cite{krizhevsky2012imagenet}. 
Throughout this study, we chose to use GraphLab's implementation of the ImageNet classifier~\cite{teterwak2014}.\footnote{GraphLab's implementation of the ImageNet classifier can be downloaded from \url{http://s3.amazonaws.com/GraphLab-Datasets/deeplearning/imagenet_model_iter45}} For each image in the dataset, we used the classifier to predict the top 5 out of 1,000 object classes that received the highest matching score by the classifier. We then counted the number in each object class (referred to as object) that appeared under each category. During the image analysis, there were cases in which certain objects were repeatedly recognized in many of the ad images. Therefore, similar to the approach of removing stop-words while processing natural language~\cite{wilbur1992automatic}, we removed objects that appeared in over 5\% of the ads (referred  to as stop-objects).

Using the method of counting the number of detected objects in each category can assist in better understanding the types of ads and objects that are used in each specific ad category. However, when using this methodology of counting objects in each category, it is still hard to understand the relationships among the various categories. Therefore, we decided to use graph visualization techniques to view the relationships among objects and categories more clearly. We defined a graph $G:=<V,E>$, where $V := Categories \cup Objects$ is a set including  all 23 ad categories and all the identified objects, and $E := \{(c,o) | c \in \mbox{Categories} \mbox{ and } o \in Objects \}$ is a set of links between a category ($c$) and an identified object ($o$), where each object appears in at least 1\% of the ads under the linked category. We then used Cytoscape~\cite{shannon2003cytoscape} and the resulting  constructed graph to visualize the connections among categories and objects. Afterwards, we used Cytoscape's GLay~\cite{su2010glay} community clustering algorithm to separate the graph into disjoint communities, and to reveal connections among the various ad categories. 

One goal of this study was to obtain an overview of the various banner ad types which exist in the datasets. To achieve this goal, we used the ImageNet classifier to convert each ad image into its corresponding vector in $\mathbb{R}^{4096}$. Next, we used the k-means++ clustering algorithm~\cite{arthur2007k} to separate the ad images into clusters. 
To use k-means++, we needed to predefine the number of clusters $k$. To make it possible to manually explore the created clusters, we needed to choose a relatively small $k$. However, we still wanted the images in each cluster to be quite similar. Therefore, to identify the most suitable $k$, we used the following simple steps: (a) for each $k \in [2,50]$, we used the k-means++ to divide the 261,752 images into $k$ disjoint clusters;\footnote{In this study, we used GraphLab's implementation of the k-means++ algorithm with maximal iteration number set to be equal to 15.} (b) for each $k \in [2,50]$, we calculated the within-cluster sum of squares (WCSS)~\cite{hartigan1979algorithm}; (c) to reduce the heuristic's result variance, we repeated steps \textit{a} and \textit{b} 50 times and calculated the mean WCSS value for each $k \in [2,50]$;  and (d) we selected the $k` \in [2,50]$ which presented the lowest mean WCSS value.

We divided the data into $k`$ clusters using k-means++. Next, we calculated the most common categories, as well as the most common objects in each cluster. We also calculated the average width and the average height of images in the cluster.
Afterwards, we randomly selected 50 images from each cluster. Then, we manually reviewed each one of these images to get a sense of the type of ads that belonged to each cluster. 

\subsection{Auto-Ads Dataset}
\label{method:auto_ads} 

One of the main goals of this study was to validate that deep-learning algorithms, such as a deep convolutional neural network, can be utilized to predict an ad's success in terms of CTR. 

By analyzing a sample of ad images in the all-ads dataset, we discovered that images from different categories tend to have different CTRs. Therefore, to achieve the goal of constructing CTR prediction models, we chose to focus on ad images in a specific category, i.e., the auto-ads dataset. Using the images in the auto-ads dataset, we constructed regression models for predicting image-ad CTRs by initiating the following steps: (a) to use only ad images that had a valid CTR, we removed from the auto-ads dataset images that were used less than 5,000 times, and had a highly exceptional CTR of over 0.2;  (b) we used the ImageNet classifier to transfer each ad image into its corresponding vector in $\mathbb{R}^{4096}$; (c)  we used Linear Regression, Random-Forest, and Boosted-Tree algorithms which are available in GraphLab's regression toolkit\footnote{\url{https://dato.com/products/create/docs/graphlab.toolkits.regression.html}} to construct prediction models that used each image's corresponding vector, running the set of regression modules with their default values except for the Boosted-Tree algorithm we set the max iterations number to be equal to 100, and for the Random-Forest algorithm we set the number of trees to be equal to 100; and (d) we evaluated the constructed regression models' performances by calculating each model's root-mean-square error (RMSE) value.

One of the factors that can influence ad performance is the ad's dimensions. In order to better understand this influence, we calculated the Pearson correlations between the ad images' \textit{width} and the ad images' CTR, and between the ad images' \textit{height} and the images' CTR. Additionally, we calculated the Pearson correlation between the ad images' \textit{pixel number}, i.e., the multiplication of each image's width and height, and the ad images' CTR. Moreover, to better grasp the influence of the ad images' dimensions and the ad images' CTR, we repeated the described-above steps, \textit{a} to \textit{d}, twice: First, we constructed regression modules using only the image width and height as features to construct the models. Second, we constructed the regression modules described above using as features each image's width and height, as well as the image's corresponding vector (this set of features is referred as the All-Features set).

\section{Results}
\label{sec:results}
In the following subsections, we present the results obtained using the algorithms and
methods described in Section~\ref{sec:methods}.  The results consist of two parts: First, in Section~\ref{sec:all_results}, we present the results of analyzing the all-ads dataset according to the methods we described in Section~\ref{method:all_ads}.  Second, in Section~\ref{sec:auto_results}, we present the results
of analyzing the auto-ads dataset according to the methods described in Section~\ref{method:auto_ads}.

\subsection{All-Ads Dataset Results} 
\label{sec:all_results}
For each image-ad category, we calculated the most common objects recognized in the ad images within the category. During our analysis, we detected the following stop-objects: (1) book jacket, dust cover (23.82\%);  (2) scoreboard (20.95\%); (3) packet (18.03\%); (4) screwdriver (15.25\%); (5) web site (15.05\%);  (6) digital clock (14.48\%); (7) street sign (10.72\%); (8) comic book (10.57\%); (10) rule, ruler (9.73\%);  (11) carpenter's kit (9.7\%);  (12) band aid (9.61\%); (13) ballpoint pen (8.42\%); (14) envelope (8.4\%); (15) ski (7.95\%); (16) menu (6.38\%); (17) rubber eraser (6.53 \%); and (18) t-shirt (5.66\%). These objects were recognized in more than 5\% of the ad images' top-5 objects. Therefore, we removed these objects.
Table~\ref{tab:category_objects} presents the most common objects that appear in each category after the removal of the stop-objects.

\begin{table}
            \caption{Most Common Identified Objects in Each Ad Category}
        \begin{center}
		\includegraphics[scale=0.9]{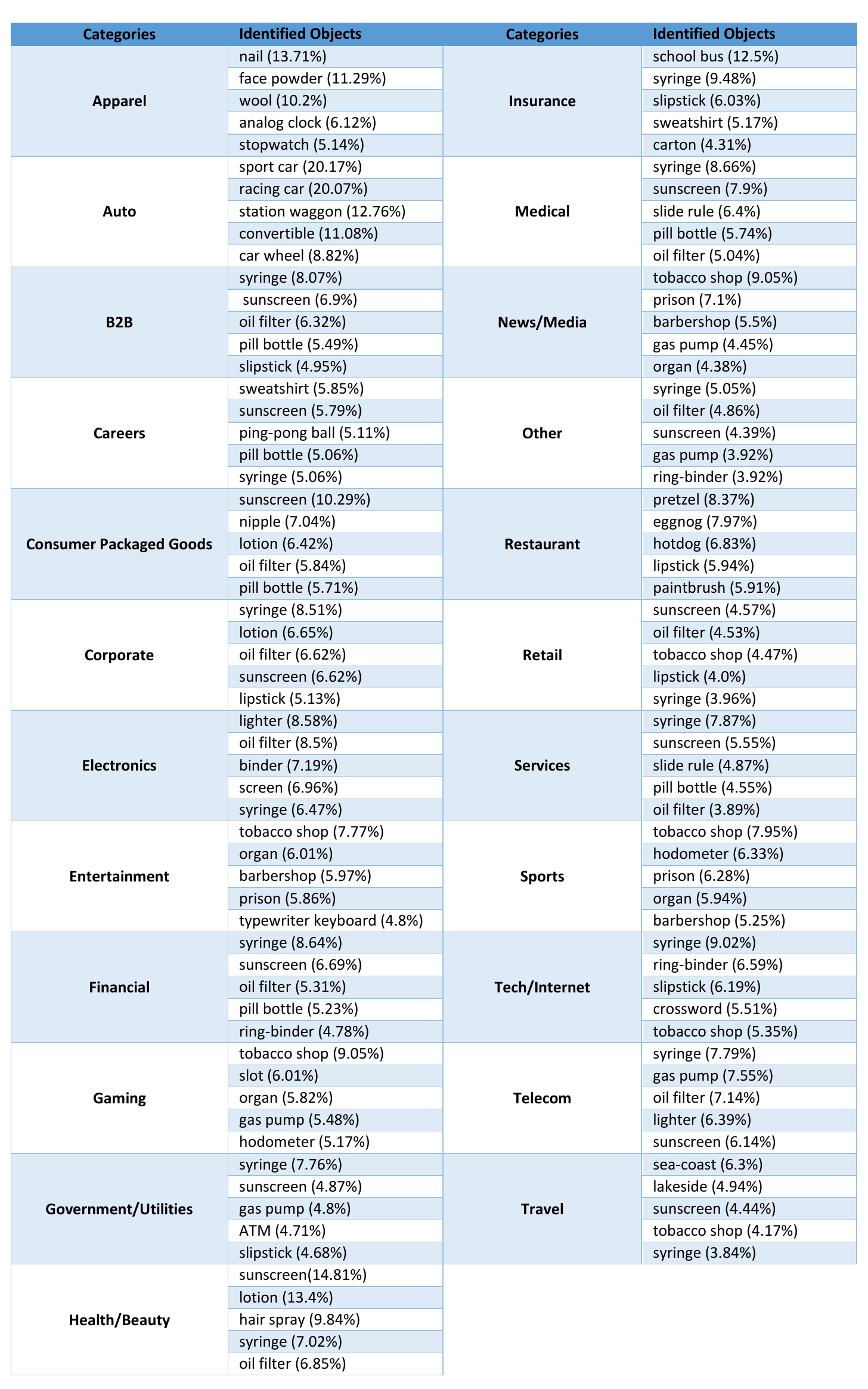}
        \end{center}
        \label{tab:category_objects}
\end{table}

\begin{figure*}
\centering
\includegraphics[scale=0.7]{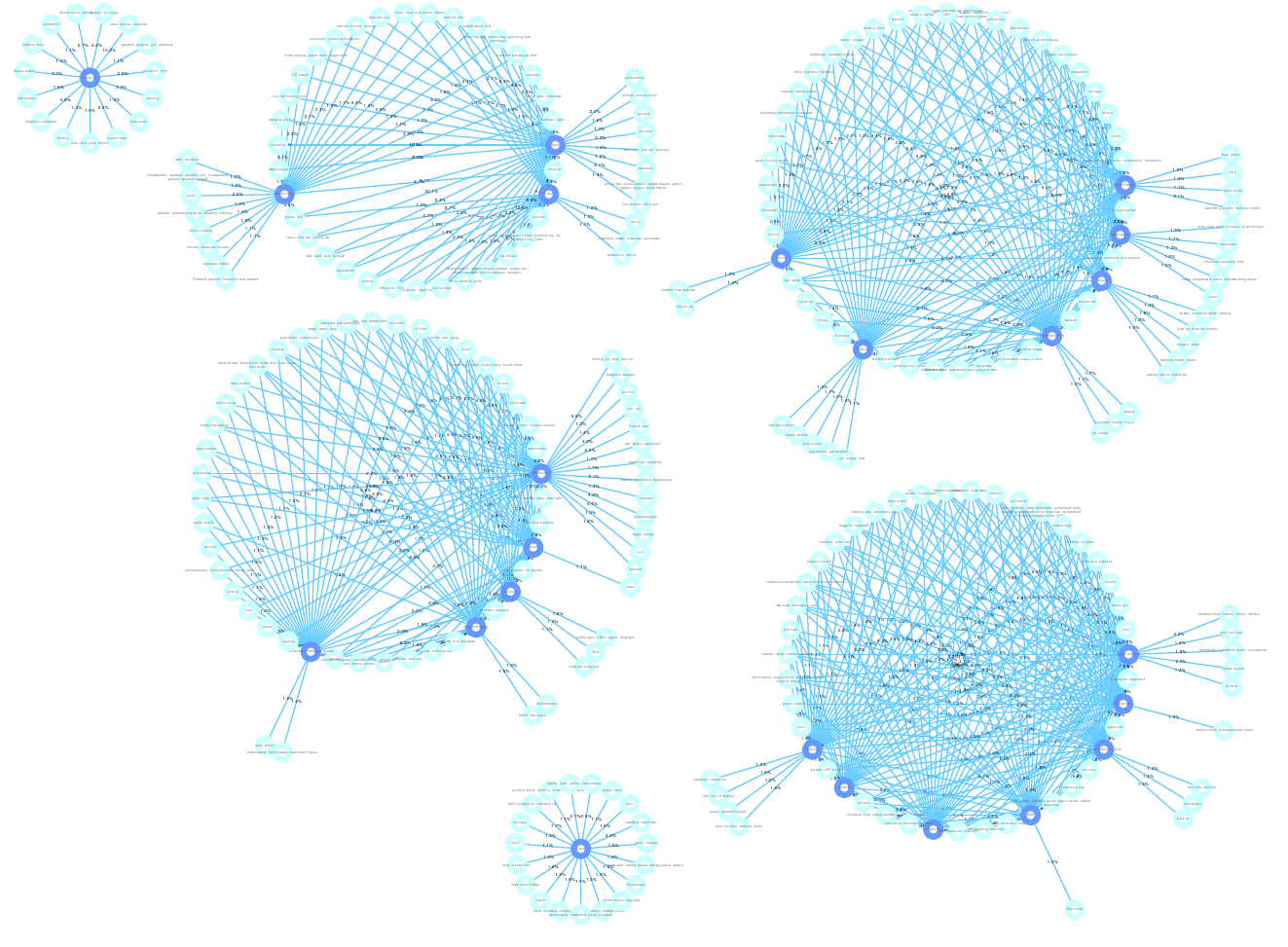}
\caption{Categories and objects graph. Category vertices are blue, and identified objects are cyan.  Each link's label contains the percentage of image ads in which each object was recognized by the image-processing algorithm in each category.  Both the links and the vertices labels are visible by zooming into the
graph.}
\label{fig:graph}
\end{figure*}

\begin{figure}
\centering
\includegraphics[scale=0.7]{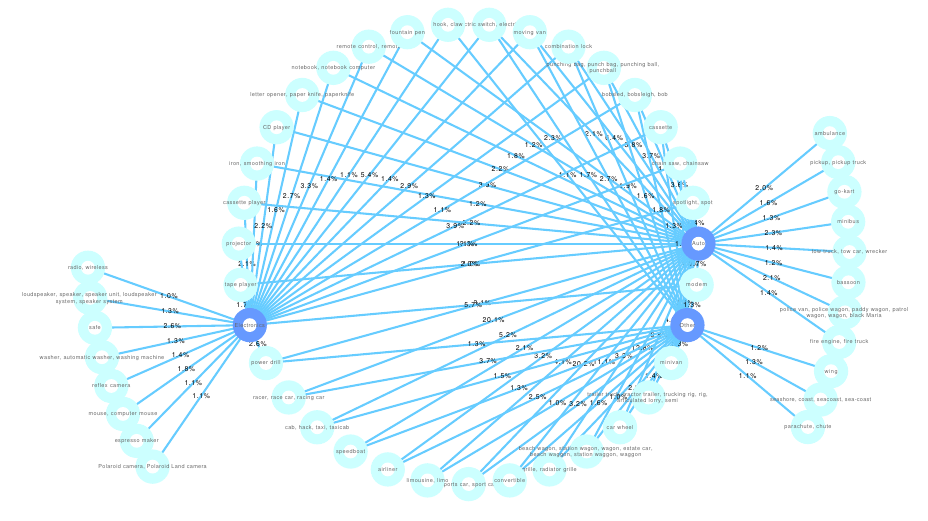}
\caption{Zoom in on the Auto, Electronics, and Other categories community.}
\label{fig:categories_zoom}
\end{figure}

Next, as described in Section~\ref{method:all_ads}, we created a graph of connections between categories and objects. The constructed graph contains 280 vertices and 1,772 links. However, due to the density of the graph, it is challenging to understand the connections among the categories and the objects. Therefore, we utilized the  GLay community detection algorithm to split the cluster into 6 communities. Figure~\ref{fig:graph} presents the six detected communities in which category vertices are marked in blue, while identified objects are marked in cyan. In addition, Figure~\ref{fig:categories_zoom} presents a zoom on the community that included the Auto, Electronics, and Other categories.

\begin{figure}
\centering
\includegraphics[scale=0.5]{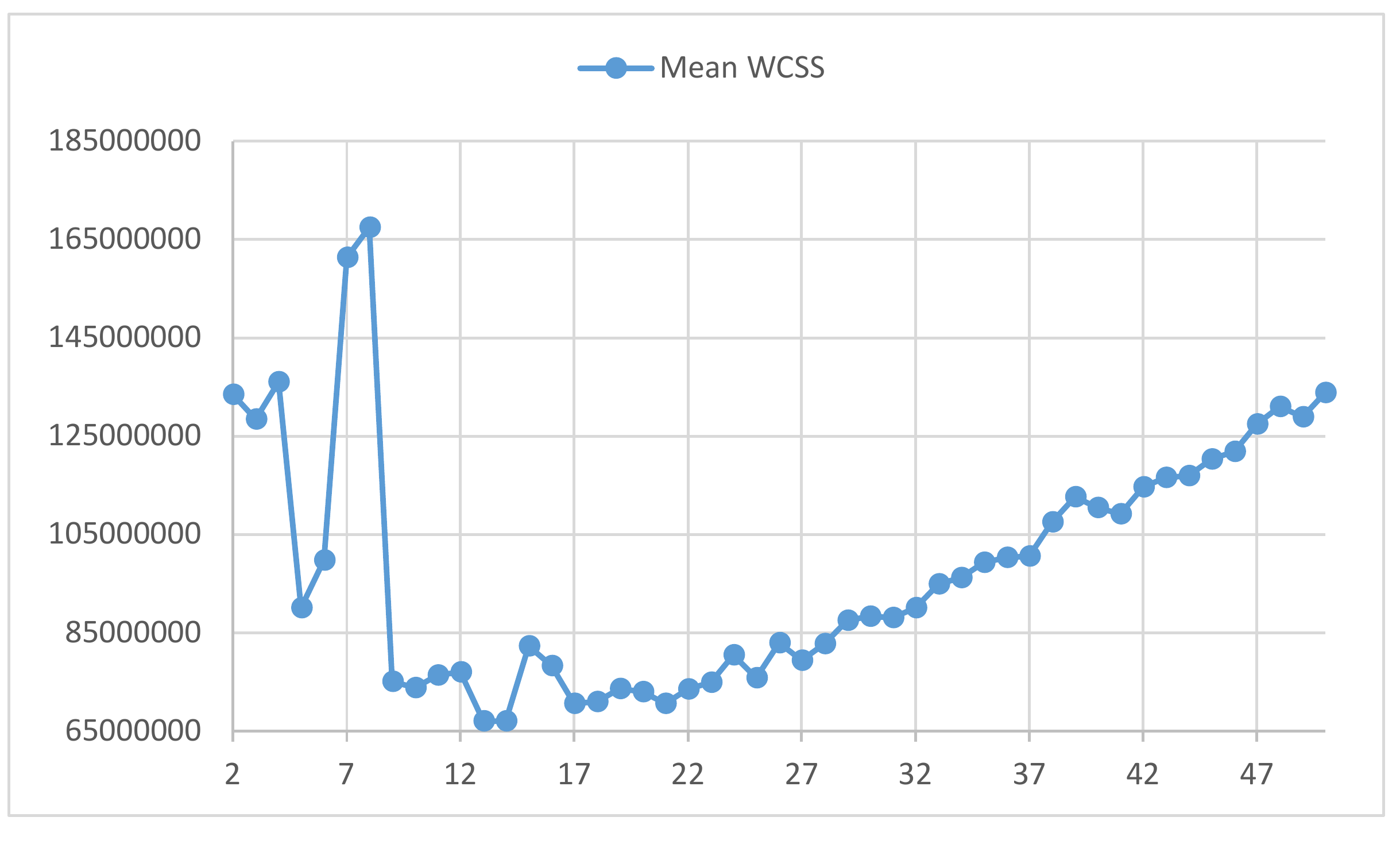}
\caption{Mean WCSS values for various $k \in [2,50]$ values. It can be observed that $k=14$ presents the minimal mean WCSS value. }
\label{fig:kmeans}
\end{figure}

Using the ImageNet classifier, we converted each image into its corresponding vector, and we used k-means++ to divide the images into $k \in [2,50]$ disjoint clusters according to the method described in Section~\ref{method:all_ads}. We discovered that using k-means++ with $k=14$ presented the minimal mean WCSS (see Figure~\ref{fig:kmeans}). Therefore, we separated the all-ads dataset into 14 clusters and analyzed each cluster. The results of these analyses are presented in Table~\ref{tab:kmeans}.

       \begin{table}
       \caption{Image Clusters using K-means++ (K=14)}
       \centering
			\includegraphics[scale=0.8]{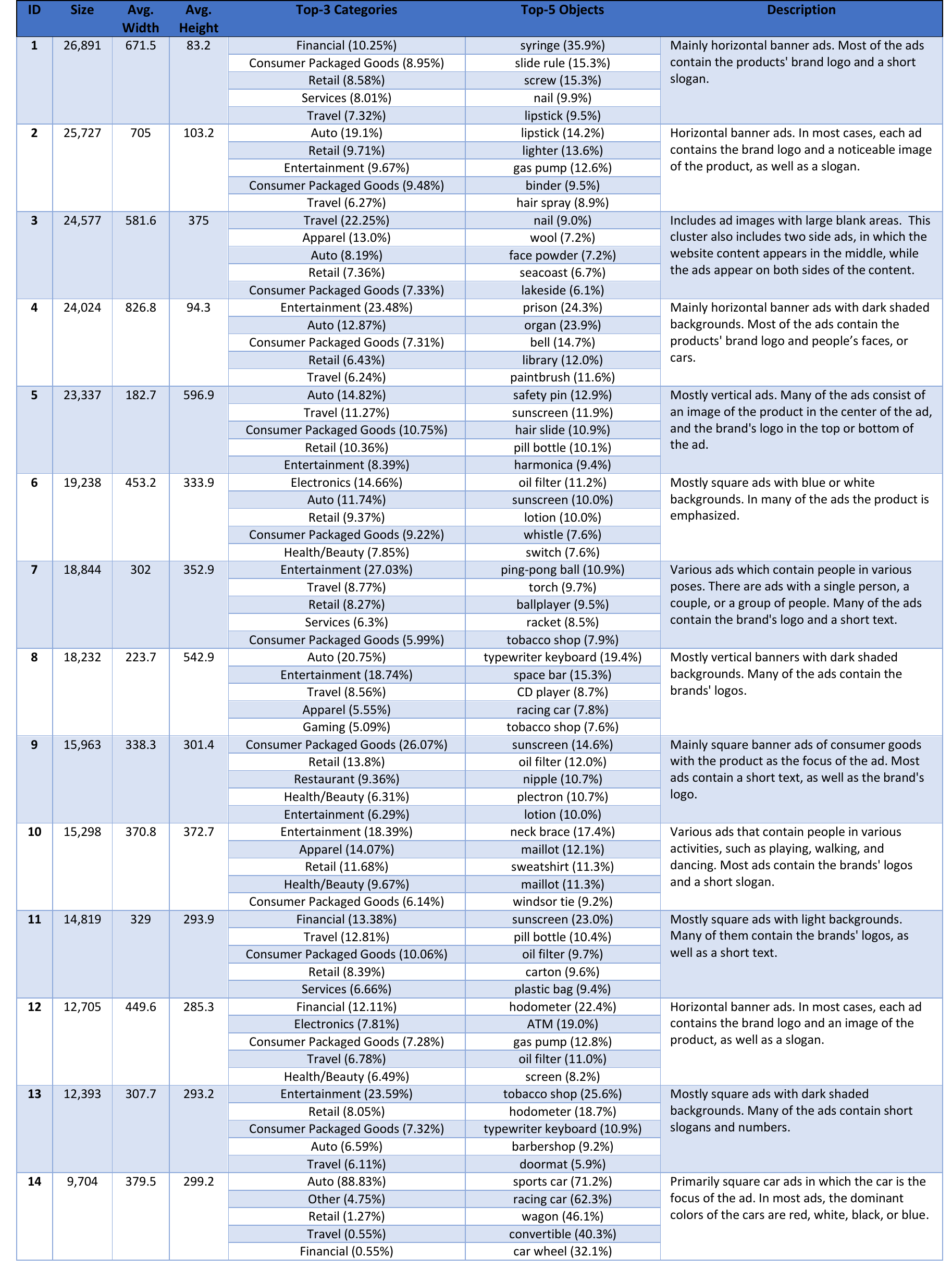}

        \label{tab:kmeans}
        \end{table}

\subsection{Auto-Ads Dataset Results} 
\label{sec:auto_results}
As a result of filtering out ad images that were presented fewer than 5,000 times and had a remarkably high CTR of over 0.2, we
were left with 12,341 ad images. Next, we calculated the following Pearson correlations between the ad images' CTR and:  (a) the ad images' \textit{width} ($r = 0.1$); (b) the ad images' \textit{height} ($r = 0.147$); and (c) the ad images' \textit{pixels number} ($r=0.237$).

Afterwards,  as described in Section~\ref{method:auto_ads}, using the 12,341 ad images, we constructed three regression models using three regression algorithms, and three set of features. Table~\ref{tab:reg} presents the RMSE of the constructed regression models.

\begin{table}[ht]
            \caption{Click-Through-Rate Prediction Results}
\centering
	\includegraphics[scale=0.95]{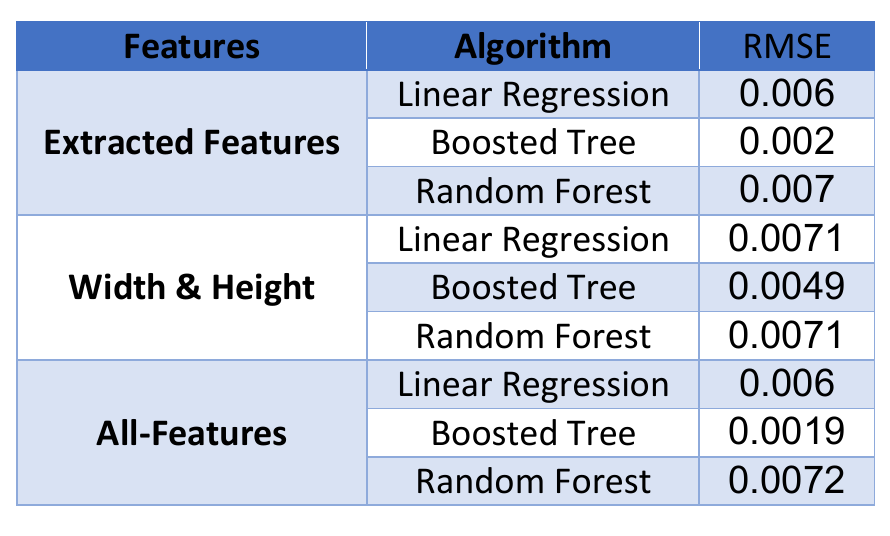}

            \label{tab:reg}
\end{table}
\section{Discussion}
\label{sec:diss}
The algorithms and methods presented throughout this study, which were evaluated on the all-ads dataset and on the auto-ads dataset revealed interesting insights that we will explore in this section. 

First, by recognizing the most common objects that appeared in each ad category and are presented in Table~\ref{tab:category_objects}, we can get a general sense of what types of ads appear in each category. 
For example, we observe that 20.17\% of ads under the Auto category present sports cars, while many ads under the Health/Beauty category present hair spray or various lotions. In addition, we can notice that the technique of removing stop-objects made Table~\ref{tab:category_objects} more readable. However, several objects, such as 
"syringe" and "sunscreen," still appeared as common objects in many of the categories. By manually inspecting over 1,000 randomly selected ads, we did not observe that these objects actually commonly appeared in the ads. Therefore, we believe that the inclusion of these objects is due to false-positive detection of the objects in many of the images. Consequently, to improve the results presented in Table~\ref{tab:category_objects}, we need to develop better stop-object detection mechanisms. We hope to explore this  research direction in a future study.

Second, as can be observed in Figures~\ref{fig:graph} and~\ref{fig:categories_zoom}, by visualizing the graph of links among  categories and recognized objects, we can easily examine which objects appear in several categories. For example, we can observe that objects specified as CD player and remote control commonly appear in more than one category. 
Moreover, we can easily observe which objects are more directly related to a specific category. For example, we can notice that steel bridges and piers commonly appear in ads under the Travel category.  This visualization technique can help in quickly understanding the main objects that appear in each category. However, this method has two main disadvantages: One disadvantage is that in the presented visualization, some of the links were removed as a result of the community detection algorithm. Moreover, we only present objects that appeared in at least 1\% of the ads under the linked category. Therefore, as a result of this process, some links that may be interesting were filtered out from the constructed graph. The other disadvantage is that the ImageNet classifier is trained to detect only 1,000 different predefined objects. Therefore, if there are other types of objects that are common in image ads, they will not appear in the created graph.

Third, by separating the objects and categories graph into communities using community detection algorithms, it is easier
to observe the links among various ad categories that contain similar objects. For example, in Figure~\ref{fig:graph}, we can see that the Auto and the Electronics categories share many objects, such as tape payer, CD player, and electric switch. Therefore, these two categories are related to each other. On the other hand, the Consumer Packaged Goods, Telecom, Retail, Health/Beauty, and Restaurant categories form a separate community which mostly contains different types of objects. By understanding the connections among the various categories we can better understand the image corpus. In a future study, we hope to show that the connections among different ad categories can be utilized to design new creative ads that will be inspired by successful ads in different categories. 

Fourth, by dividing the ad images into 14 clusters using k-means++, and then analyzing each cluster (see Table~\ref{tab:kmeans}), we can reveal several interesting insights regarding the ads in our corpus. We can notice that even though the image ads are separated into different categories, many of the ads under different categories share similar characteristics. For example, most of the ads contain the brand logo, and many of the images contain the brand's slogan or a short text. Additionally, while viewing a sample set from each cluster, we also observed several types of ad images. For instance, we observed the following common types of ads: (a) \textit{text ads}, in which most of the ads contain mainly text, rather than images; (b) \textit{models ads}, in which images of people are the primary focus of the ad, and in many cases these people are performing various actions or standing in various poses; and (c) \textit{product ads}, in which an image of the product is  the focus of the ad, something especially  common in ads under the Auto category.
We also observed that ad images with similar width and height proportions, such as horizontal banner ads, present similar characteristics. Using our clustering, we additionally succeeded in identifying less common types of ads that appeared as part of cluster number 3. Cluster 3 contains ad images that wrap the website content from both sides. 

Lastly, according to the results presented in Section~\ref{sec:auto_results}, we found negligible positive correlation ($r=0.1$) between the images' width and the images' CTR, as well as negligible positive correlation ($r=0.147$) between images' height and the images' CTR. Additionally, we revealed weak positive correlation ($r=0.237$) between the images' pixel number and the images' CTR. As expected, these results indicate that larger image ads indeed tend to have higher CTRs. However, as can be inferred  from Table~\ref{tab:reg}, we can construct a more accurate CTR prediction model, with an RMSE as low as 0.0019,  by using both the size features of the ad as well as the 4,096 features that were extracted using the ImageNet classifier. These results indicate that deep-learning image processing algorithms can assist in predicting an image ad's success.

\section{Conclusions and Future Work }

In this study, we utilized deep-learning image processing algorithms as well as clustering algorithms to explore a large-scale categorized image corpus. We demonstrate that even though the ad-image corpus contains over 250,000 images, our algorithms make it possible to better understand the various types and layouts of ads that exist in this corpus by sampling only a small subset of corpus, which contains about 1,000 ad images (see Table~\ref{tab:kmeans}). The methods that are presented throughout this study can also be employed to investigate other categorized and even uncategorized large-scale image corpora to reveal significant patterns and insights.
Moreover, by utilizing deep-learning algorithms and extracting the common objects that appear in each category, we show that it is possible to get a sense of which types of objects appear in each category (see Table~\ref{tab:category_objects} and Figure~\ref{fig:graph}), and to even understand the connections among the various categories (see Figures~\ref{fig:graph} and~\ref{fig:categories_zoom}). We believe that a better understanding of the connections among the various ad categories and various objects can influence advertisers. They can gain a fresh perceptive on the impact of their ads and can discern which elements need to be altered, removed, or incorporated to produce more original and effective ads. For example, ad designers who are considering embedding a specific object into their ad can learn from other ads that have already embedded this object.  

According to Section~\ref{sec:auto_results} results, we can observe that regression models, which utilize features that were extracted using the ImageNet classifier, can predict the CTR of image ads with an RMSE as low as 0.0019 (see Table~\ref{tab:reg}). These models are considerably better than naive models that use only the images' dimensions to predict the ads' CTR. This type of prediction model can be instrumental in helping advertisers create more successful ads, resulting in higher traffic to their websites and subsequent increases in sales.  Moreover, it can also help them to quickly pinpoint unsuccessful ads and make changes accordingly. 

This study is the first of its kind and offers many research directions to pursue. One interesting direction is to improve the image corpus exploration methods presented throughout this study and to apply them on other types of image corpora. 
Another possible future research direction is to develop algorithms for exploring and analyzing video ads, and for predicting video ad success.  A further interesting research direction is to improve the regression models presented in this study by constructing these models with additional features, such as features that include information on the web pages in which the ads were published. 
One additional possible future research direction includes developing deep-learning algorithms specifically for optimizing the design of image ads. These algorithms can identify objects that, by adding them to an ad, will increase the ad's performance. For example, these types of algorithms can reveal if adding a black sports car to an ad is better than adding a red SUV. They could even recommend exactly which objects to embed in an image ad in order to directly increase the ad's CTR. We hope to explore this research direction in a future study.

Overall, the results presented in this study, as well as the offered future research directions, emphasize the vast potential that exists in utilizing deep-learning algorithms in the domain of online advertising. This market is growing swiftly, and the application of increasingly sophisticated analysis and predictive methods is vital.

\label{sec:conclusions}
\section{Acknowledgments}
We would like to thank Yuval Niv for
providing us with the ad image corpus for this  article.  Additionally,  we  want  to  thank  Carol Teegarden for editing and proofreading this article.

%
\bibliographystyle{abbrv}
\bibliography{deep_ads}  
\end{document}